# My camera can see through fences: A deep learning approach for image de-fencing

Sankaraganesh Jonna[1], Krishna K. Nakka[2], and Rajiv R. Sahay[1,2]

[1]School of Information Technology
[2]Department of Electrical Engineering
Indian Institute of Technology Kharagpur, Kharagpur, India
{sankar9.iitkgp, krishkanth.92, sahayiitm}@gmail.com

## Abstract

*In recent times, the availability of inexpensive image capturing devices such as smartphones/tablets has led to an exponential increase in the number of images/videos captured. However, sometimes the amateur photographer is hindered by fences in the scene which have to be removed after the image has been captured. Conventional approaches to image de-fencing suffer from inaccurate and non-robust fence detection apart from being limited to processing images of only static occluded scenes. In this paper, we propose a semi-automated de-fencing algorithm using a video of the dynamic scene. We use convolutional neural networks for detecting fence pixels. We provide qualitative as well as quantitative comparison results with existing lattice detection algorithms on the existing PSU NRT data set [1] and a proposed challenging fenced image dataset. The inverse problem of fence removal is solved using split Bregman technique assuming total variation of the de-fenced image as the regularization constraint.*

## 1. Introduction

A challenging problem faced by photographers is when the subject they want to capture is occluded by a fence. Several exhibits, museum showpieces, landmarks, are hidden behind fences and barricades due to security concerns. Image de-fencing involves the removal of fences or occlusions in such degraded photographs. Several works such as [2, 4, 6, 20, 17, 7] addressed the image inpainting problem wherein the portion of the image to be inpainted is specified by a mask manually. In contrast for the problem at hand it is difficult to manually mark all fence pixels since they are numerous and cover the entire image.

In fact in this work, we view the detection of fence pixels in an occluded observation as a machine learning problem and seek to automate the process. The challenge is to robustly isolate fences in the presence of clutter, bad illumination, perspective distortion etc. Image inpainting does not yield satisfactory results when the image contains fine textured regions that have to be filled-in [13, 19]. However, using a video captured by moving a camera relative to a fenced scene can lead to better results due to availability of additional information in the adjacent frames. Hence, we recognize that image de-fencing using a captured video involves multiple tasks such as fence detection, relative motion estimation among the observations and information fusion using them.

The authors in [13] first addressed the image de-fencing problem via inpainting of fence occlusions using Criminisi et al's method [4]. Subsequently, the the authors in [19] extended the work of [13] using a video of the scene for de-fencing, significantly improving the performance due to availability of hidden information in additional frames. The work in [19] is closest to the proposed algorithm for image de-fencing since it also uses a video of dynamic occluded scenes. Vrushali et al. [9] proposed an improved multi-frame de-fencing technique. However, the work in [9] assumed that motion between the frames is global. This assumption is invalid for more complex dynamic scenes where the motion is non global. Also, the method of [9] used an image matting technique [21] for fence detection which involves significant user interaction. We overcome these two important drawbacks of [9] in the proposed work.

Apart from the image based techniques, Jonna et al. [8] proposed multimodal approach for image de-fencing wherein they have extracted the fence masks by depth maps corresponding to the color images obtained using Kinect sensor. In a very recent work, a video de-fencing algorithm is proposed in [14]. However, Yadong et al's method [14] restricts the motion of the camera to be affine and users need training to capture the video in an appropriate manner. Importantly, the algorithm in [14] failed to distinguish between the parallax caused by fences as well as dynamic scenes in the background. A major drawback of [14] is that algorithm is limited to defencing static background scenes. Another limitation of [14] is that the authors do not provide a comprehensive evaluation of their algorithm for fence localization/detection. We address these limitations of [14] in this paper. Specifically, the proposed method can handle reasonably arbitrary camera motion such as rotation, zooming etc. during video capture. Also, we can de-fence scenes containing dynamic objects.

Although, there have been several previous attempts for removing fences from images [13, 19, 9, 15, 14, 8] the novelty in our approach is a deep learning based framework for fence detection and formulation of sparsity based optimization framework to fuse data from multiple frames of the input video in order to fill-in fence pixels. Importantly, unlike previous works [14] for the task of fence detection, we provide qualitative and quantitative comparison results with existing lattice detection algorithms [18] on the PSU NRT data set [1] as well as on a proposed fenced image dataset containing several challenging scenarios.

## 2. Proposed methodology

We propose to model the image affected by fences as

$$\mathbf{y}_m = \mathbf{O}_m \mathbf{W}_m \mathbf{x} + \mathbf{n}_m \quad (1)$$

where $\mathbf{y}_m$ represents the $m^{\text{th}}$ frame of the video, $\mathbf{O}_m$ is the binary fence mask corresponding to $m^{\text{th}}$ frame, $\mathbf{W}_m$ is the warp matrix, $\mathbf{x}$ denotes the de-fenced image and $\mathbf{n}_m$ is the noise assumed as Gaussian.

The overall flow of the image de-fencing approach is shown in Fig. 1. We propose to solve the inverse problem corresponding to Eq. (1) by addressing each of the three sub-problems of fence detection, optical flow estimation and data fusion.

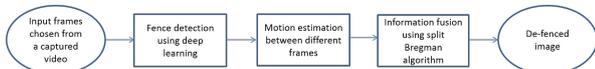

Figure 1. Workflow of proposed deep learning based image defencing algorithm.

### 2.1. Fence detection using convolutional neural network

Convolutional neural networks were originally proposed by LeCun [11] and have proven their utility for several challenging real-world machine learning problems [10, 12]. According to LeCun [11] CNNs can be effectively trained to recognize objects directly from their images with robustness to scale, shape, angle, noise etc and have been shown to outperform classification schemes using hand crafted features [10]. Hence, we are motivated to investigate the utility of CNNs for the task of fence detection which we pose as a machine learning problem in this work. The overall architecture of the proposed convolutional neural network is described in Fig. 2. It contains five layers- two convolutional and two pooling layers followed by a single fully connected output layer. The output maps from layer 4 are concatenated to form a single vector while training and fed to the next layer. The final output layer which contains two neurons corresponding to each class, are fully connected by weights with the previous layer. Akin to the neurons in the convolutional layer, the responses of the output layer neurons are also modulated by the non linear sigmoid function to produce the resultant score for each class.

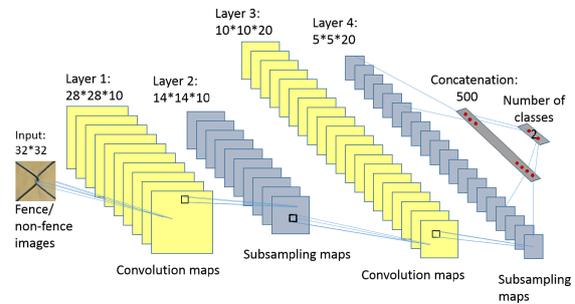

Figure 2. Architecture of the proposed CNN model used for fence/non-fence texel classification.

The base dataset for training of the proposed CNN architecture consists of 4000 positive and 8000 negative examples of fence texels, a sample of which is shown in Fig. 3 (a), (b). The positive images have tightly cropped fence joints and negative examples contain images of non-joints. To improve classification performance, we have augmented our dataset of fence texels by cropping and flipping. The input image is cropped to $26 \times 26$ pixels at four corners and later resized to $32 \times 32$ pixels. Flipping is done along Y-axes to augment the data by 5 times.

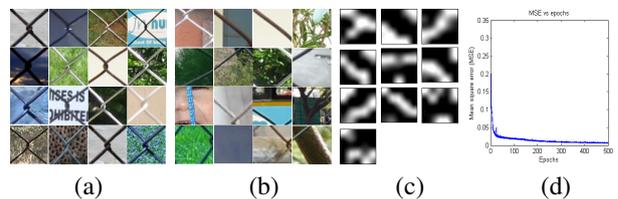

Figure 3. (a) Sample positive examples of fence texels. (b) Sample non-fence texels. (c) Kernel weights of layer 1. (d) MSE vs epochs plot.

The weights of the proposed CNN are trained by the conventional back-propagation method using the package in [16]. The total number of learnable parameters in the proposed CNN architecture is 6282. We have chosen batch size as 50 and constant learning rate $\alpha$=0.5 throughout all the layers. The network is trained for 500 epochs.

For the sake of visualization, we show the kernels from first convolutional layer of the trained CNN in Fig. 3 (c). We can observe that the filters from layer 1 have active weights along the fence gradient directions. The variation of mean square error (MSE) versus epochs during the training phase is plotted in Fig. 3 (d) and the final MSE achieved during training is 0.02%.

The trained CNN model is used to classify the fence pixels in the entire test image by moving a detector window of size $32 \times 32$ pixels with a stride of 4 pixels. The centroid of the positively detected fence texels are identified as the estimated joint locations which are connected throughout the entire input image to obtain the final fence mask which corresponds to $\mathbf{O}_m$ in Eq. 1.

### 2.2. Motion estimation

We capture a short video clip by moving a camera relative to the fenced scene and use few frames from it to de-fence the reference image. It is natural to assume that pixels occluded in the reference image are uncovered in the additional frames of the video of the landscape which can be static or dynamic. We use the method of [3] to obtain the optical flow and hence the warping matrix $\mathbf{W}_m$ in Eq. (1).

### 2.3. Data fusion using sparsity based regularization

We solve for the de-fenced image as an inverse problem. We assume total variation (TV) [17] of the de-fenced image as the regularization constraint. The de-fenced image is the solution of the following constrained optimization problem

$$\arg\min_{\mathbf{x}} \frac{1}{2} \sum_{m=1}^{p} \| \mathbf{y}_m - \mathbf{O}_m \mathbf{W}_m \mathbf{x} \|_2^2 + \mu \| \mathbf{d} \|_1 \ s.t. \ \mathbf{d} = \nabla \mathbf{x}$$

where $p$ is the number of frames chosen from the video and $\mu$ is the regularization parameter. The above optimization framework is a combination of both $l1$ and $l2$ terms and hence difficult to solve. We employ the split Bregman iterative framework described in [5] to solve the above problem. We use an alternative unconstrained formulation as

$$\arg\min_{\mathbf{x}} \frac{1}{2} \sum_{m=1}^{p} \| \mathbf{y}_m - \mathbf{O}_m \mathbf{W}_m \mathbf{x} \|_2^2 + \mu \| \mathbf{d} \|_1 \quad (2)$$
$$+ \frac{\lambda}{2} \| \mathbf{d} - \nabla \mathbf{x} \|_2^2$$

where $\lambda$ is the shrinkage parameter. The iterates to solve the above equation are as

$$[\mathbf{x}^{k+1}, \mathbf{d}^{k+1}] = \arg\min_{\mathbf{x},\mathbf{d}} \frac{1}{2} \sum_{m=1}^{p} \| \mathbf{y}_m - \mathbf{O}_m \mathbf{W}_m \mathbf{x}^k \|_2^2 \quad (3)$$
$$+ \mu \| \mathbf{d}^k \|_1 + \frac{\lambda}{2} \| \mathbf{d}^k - \nabla \mathbf{x}^k + \mathbf{b}^k \|_2^2$$

$$\mathbf{b}^{k+1} = \nabla \mathbf{x}^{k+1} + \mathbf{b}^k - \mathbf{d}^{k+1}$$

We can now split the above problem into two sub-problems as

**Sub Problem 1:**

$$[\mathbf{x}^{k+1}] = \arg\min_{\mathbf{x}} \frac{1}{2} \sum_{m=1}^{p} \| \mathbf{y}_m - \mathbf{O}_m \mathbf{W}_m \mathbf{x}^k \|_2^2$$
$$+ \frac{\lambda}{2} \| \mathbf{d}^k - \nabla \mathbf{x}^k + \mathbf{b}^k \|_2^2$$

This sub-problem is solved by a gradient descent method.

**Sub Problem 2:**

$$[\mathbf{d}^{k+1}] = \arg\min_{\mathbf{d}} \mu \| \mathbf{d}^k \|_1 + \frac{\lambda}{2} \| \mathbf{d}^k - \nabla \mathbf{x}^{k+1} + \mathbf{b}^k \|_2^2$$

The above sub-problem can be solved by applying the shrinkage operator as follows

$$\mathbf{d}^{k+1} = shrink(\nabla \mathbf{x}^{k+1} + \mathbf{b}^k, \frac{\lambda}{\mu})$$

$$\mathbf{d}^{k+1} = \frac{\nabla \mathbf{x}^{k+1} + \mathbf{b}^k}{| \nabla \mathbf{x}^{k+1} + \mathbf{b}^k |} * max(| \nabla \mathbf{x}^{k+1} + \mathbf{b}^k | - \frac{\lambda}{\mu}, 0)$$

The update for $\mathbf{b}$ is as $\mathbf{b}^{k+1} = \nabla \mathbf{x}^{k+1} + \mathbf{b}^k - \mathbf{d}^{k+1}$. We tune the parameters $\mu, \lambda$ to obtain the best estimate of the de-fenced image.

## 3. Results and Discussions

Initially, we report experimental results of the proposed deep learning algorithm for fence detection on the two datasets considered in this work. Subsequently, we also report the de-fencing results obtained using our sparsity based optimization framework. We tuned the parameters $\mu$ and $\lambda$ and found empirically that $\mu = 0.01$ and $\lambda = 1e-5$ yielded the best results. Note that we used just 3 to 4 frames from the captured video to obtain all our results.

### 3.1. Fence Detection

For evaluating our algorithm for fence detection, we have collected a dataset consisting of 200 real-world images/videos under several challenging scenarios such as clutter, bad lighting, viewpoint distortion etc. We have also evaluated our algorithm on 32 images from the PSU NRT database [1] containing fences. In Fig. 4 we provide a qualitative comparison of the performance of the method in [18] with the proposed algorithm. Images in the first row are from our dataset whereas those in the second row are from PSU NRT database. We observe that the method of [18] partially detected fence texels shown in Figs. 4 (a), (e) whereas our algorithm detected fence pixels faithfully in Figs. 4 (b), (f), respectively. Importantly, the algorithm in [18] failed to detect many fence texels shown in Figs. 4 (c), (g). However, our algorithm demonstrate superior performance by detecting all fence pixels robustly shown in Figs. 4 (d), (h).

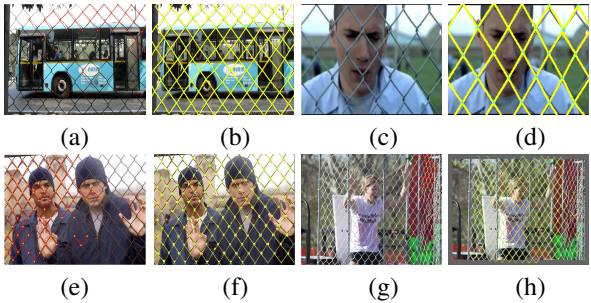

Figure 4. Some results for fence detection: First column shows partially detected fences using [18]. Second column: Corresponding fence detection results using the proposed algorithm. Third column: Method of [18] failed to detect many fence texels. Fourth column: Fence detection using the proposed algorithm.

A summary of the quantitative evaluation of the fence detection method of [18] and the CNN based proposed algorithm is given in Table 1.

Table 1. Quantitative evaluation of fence detection

| Method | Our Database | | | NRT Database [1] | | |
|---|---|---|---|---|---|---|
| | Precision | Recall | F-measure | Precision | Recall | F-measure |
| Park et al. [18] | 0.94 | 0.26 | 0.41 | 0.95 | 0.46 | 0.62 |
| **Proposed method** | 0.84 | 0.96 | **0.89** | 0.94 | 0.96 | **0.95** |

### 3.2. Image de-fencing

Initially, we report results of the proposed method on synthetic data. The original image is shifted with different pixel displacements of $(-5, -5)$, $(2, 2)$ and $(10, 10)$ pixels to obtain four different frames and then super-imposed by a synthetically generated fence. In Fig. 5 (a) we show the motion vectors between first and fourth observations overlaid on the first frame. In Fig. 5 (b) we show the de-fenced image obtained using the proposed technique when the relative motion between the frames are wrongly given as $(-3, -3)$, $(4, 4)$ and $(8, 8)$. Observe that undesired artifacts appear and the fence is not completely removed in Fig. 5 (b). The de-fenced image obtained by the proposed method using the correct relative motion is shown in Fig. 5 (c). Note that there are hardly any artifacts and the fence has been successfully removed. The de-fenced image shown in Fig. 5 (c) was found to have a PSNR of 39.84dB and SSIM of 0.99. This synthetic experiment shows the importance of each module (Fig. 1) in our de-fencing approach

As a challenging example, we next conducted experiments on a video taken from the famous 'Prison Break' TV shows. In Fig. 6 (a) we show the first frame overlaid with optical flow vectors between it and third observation. In Figs. 4 (c), (d) we show the fence detection results for this case obtained using [18] and proposed deep learning algorithm, respectively. The estimated de-fenced image obtained using our algorithm is shown in Fig. 6 (b).

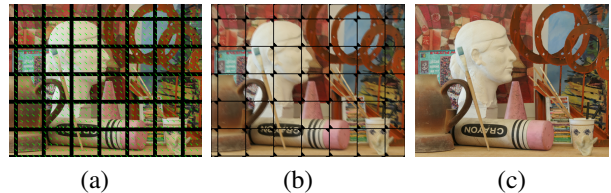

Figure 5. (a) Frame chosen from a video. (b) De-fenced image using proposed algorithm with wrongly estimated motion between frames. (c) De-fenced image obtained using proposed algorithm with accurately estimated motion among frames.

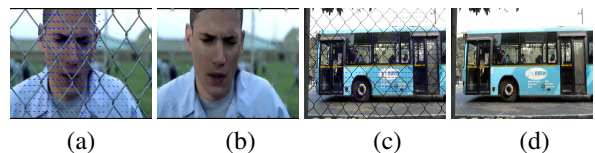

Figure 6. (a), (c) are frames chosen from videos. (b), (d) are corresponding de-fenced images obtained using proposed algorithm.

We experimented with one more challenging real-world traffic video. The motion vectors between first and third frames obtained using [3] are shown overlaid on first frame in Fig. 6 (c). In Figs. 4 (a), (b) we show the fence detection results for this case obtained using [18] and proposed CNN-based algorithm respectively. Note that the proposed CNN-based approach detected the fence pixels in dark regions such as over the two doors and tyre of the wheel of the bus. The final de-fenced image obtained using our optimization framework is shown in Fig. 6 (d).

Since the method in [14] is unable to de-fence dynamic scenes we choose to provide comparison results with an existing video-based image de-fencing technique in [19] only. In Fig. 7 (a), we show a fenced frame overlaid with optical flow vectors between it and third observation taken from a video used in [19]. The de-fenced image obtained in [19] is shown in Fig. 7 (b). We can clearly observe several artifacts on the top of the head and on the T-shirt of the person facing the camera. Also, the lips have not been reconstructed properly. In the close-up shown in the inset of Fig. 7 (b), we can see that there are still some artifacts near the lip region. We used the same example video and extracted three frames from it. The de-fenced image obtained using proposed algorithm on the same three frames from the video in [19] is shown in Fig. 7 (c). We can clearly observe the superiority of proposed algorithm over [19].

The proposed de-fencing algorithm failed to restore the occluded images if there are any errors in motion estimation between the frames. In Fig. 8 (a) we show the image chosen from a tennis video and the corresponding de-fenced image obtained using proposed algorithm is shown in Fig. 8 (b).

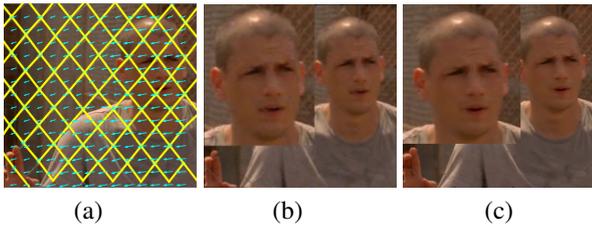

(a) (b) (c)

Figure 7. (a) One frame chosen from a video. (b) De-fenced image using the video-based algorithm in [19]. (c) De-fenced image using proposed method.

In the close-up shown in the inset of Fig. 8 (b), we can see that there are some artifacts near the legs and tennis racket regions due to the local and complex motion of player.

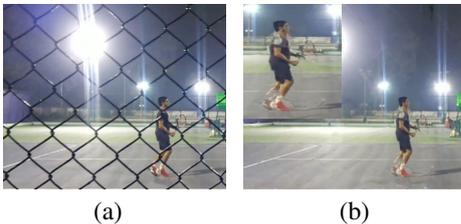

(a) (b)

Figure 8. Failure case: (a) One frame chosen from a video. (b) De-fenced image obtained using proposed algorithm.

## 4. Conclusions

We proposed an approach for de-fencing an image using multiple frames from a video captured by moving a camera relative to a static or dynamic scene. Our approach for image de-fencing necessitates the solution of three sub-problems: (i) automatic detection of fences or occlusions (ii) accurate estimation of relative motion between the frames (iii) data fusion to fill-in occluded pixels in the reference image. We have proposed an automated CNN-based framework for the first sub-problem and to validate its accuracy we present a new challenging fenced image dataset consisting of 200 images. The split Bregman optimization approach was to obtain the de-fenced image. Our results for both synthetic and real-world data show the superiority of the proposed algorithm over the state-of-the-art techniques.